\documentclass{ecai} 
\usepackage{latexsym}
\usepackage{amssymb}
\usepackage{amsmath}
\usepackage{amsthm}
\usepackage{booktabs}
\usepackage{enumitem}
\usepackage{graphicx}
\usepackage{color}
\usepackage{microtype}
\usepackage{graphicx}
\usepackage{subfigure}
\usepackage{booktabs} % for professional tables
\usepackage[table]{xcolor}
\usepackage{hyperref}
\usepackage{algorithm}
\usepackage{algorithmic}

\DeclareMathOperator*{\argmin}{arg\,min}
\DeclareMathOperator*{\arginf}{arg\,inf}

\newtheorem{theorem}{Theorem}
\newtheorem{lemma}[theorem]{Lemma}

\newtheorem{definition}{Definition}

\newcommand{\BibTeX}{B\kern-.05em{\sc i\kern-.025em b}\kern-.08em\TeX}

\begin{document}

%%%%%%%%%%%%%%%%%%%%%%%%%%%%%%%%%%%%%%%%%%%%%%%%%%%%%%%%%%%%%%%%%%%%%%%%

\begin{frontmatter}

%%% Use this command to specify your submission number.
%%% In doubleblind mode, it will be printed on the first page.

\paperid{123} 

%%% Use this command to specify the title of your paper.

\title{Global Optimisation of Black-Box Functions with Generative Models in the Wasserstein Space}

%%% Use this combinations of commands to specify all authors of your 
%%% paper. Use \fnms{} and \snm{} to indicate everyone's first names 
%%% and surname. This will help the publisher with indexing the 
%%% proceedings. Please use a reasonable approximation in case your 
%%% name does not neatly split into "first names" and "surname".
%%% Specifying your ORCID digital identifier is optional. 
%%% Use the \thanks{} command to indicate one or more corresponding 
%%% authors and their email address(es). If so desired, you can specify
%%% author contributions using the \footnote{} command.

\author[A]{\fnms{Tigran}~\snm{Ramazyan}\orcid{0000-0002-8770-0637}\thanks{Corresponding Author. Email: \href{mailto:tramazyan@hse.ru}{tramazyan@hse.ru}.\\Original code: \url{https://github.com/ramazyant/wu-go}\\Python package: \url{https://github.com/hse-cs/waggon}}}
\author[A]{\fnms{Mikhail}~\snm{Hushchyn}\orcid{0000-0002-8894-6292}}
\author[A]{\fnms{Denis}~\snm{Derkach}\orcid{0000-0001-5871-0628}}

\address[A]{HSE University, Moscow, Russia}

%%% Use this environment to include an abstract of your paper.

\begin{abstract}We propose a new uncertainty estimator for gradient-free optimisation of black-box simulators using deep generative surrogate models. Optimisation of these simulators is especially challenging for stochastic simulators and higher dimensions. To address these issues, we utilise a deep generative surrogate approach to model the black box response for the entire parameter space. We then leverage this knowledge to estimate the proposed uncertainty based on the Wasserstein distance - the Wasserstein uncertainty. This approach is employed in a posterior agnostic gradient-free optimisation algorithm that minimises regret over the entire parameter space. A series of tests were conducted to demonstrate that our method is more robust to the shape of both the black box function and the stochastic response of the black box than state-of-the-art methods, such as efficient global optimisation with a deep Gaussian process surrogate.\end{abstract}

\end{frontmatter}

% We propose a new uncertainty estimator based on Wasserstein balls. We pair it with a deep generative model for gradient-free optimisation of black-box simulators with non-differentiable objectives. Optimisation of these simulators is especially challenging for stochastic simulators and higher dimensions. To address these issues, we utilise a deep generative surrogate approach to model the black box response for the entire parameter space. We then leverage this knowledge to estimate the proposed uncertainty. This approach is employed in a posterior agnostic gradient-free optimisation algorithm that minimises regret over the entire parameter space. A series of tests were conducted to demonstrate that our method is more robust to the shape of both the black box function and the stochastic response of the black box than state-of-the-art methods that utilise uncertainty for optimisation, such as efficient global optimisation with a deep Gaussian process surrogate.

%%%%%%%%%%%%%%%%%%%%%%%%%%%%%%%%%%%%%%%%%%%%%%%%%%%%%%%%%%%%%%%%%%%%%%%%

\section{Introduction}
\label{intro}

Simulation of real-world experiments is key to scientific discoveries and engineering solutions. Such techniques use parameters describing configurations and architectures of the system. A common challenge is to find the optimal configuration for some objective, e.g., such that it maximises efficiency or minimises costs and overhead~\cite{DORIGO2023100085}.

To generalise the problem, we examine the case of stochastic simulators. A single call of such simulators is often computationally expensive, which is the case for many domains, especially when the simulator is Monte Carlo-based. For instance simulators modelling particle scatterings~\cite{lgso56}, molecular dynamics~\cite{lgso1}, and radiation transport~\cite{lgso4}. The problem is also often referred to as design optimisation~\cite{lgso44}.

In this case, the simulator is treated as a black box experiment. This means that observations come from an unknown and likely difficult-to-estimate function. Using a surrogate model for black-box optimisation (BBO) is an established technique~\cite{cozad_surrogate}, as BBO has a rich history of both gradient and gradient-free methods ~\cite{bbo20}, most of which come from tackling problems that arise in physics, chemistry and engineering. 

Although gradient-based optimisation has proven to work well for differentiable objective functions, e.g. ~\cite{DBLP:journals/corr/ChangUTT16, DBLP:journals/corr/DegraveHDW16, NEURIPS2018_842424a1}, in practice nondifferentiable objective functions appear often in applications~\cite{Aehle_2023}. For instance, when one addresses parameter optimisation of a system represented by a Monte Carlo simulator which provides data from an intractable likelihood~\cite{JMLR:v17:15-017}. In addition to that, simulation samples used as input to train machine learning models usually consist of several samples representing disconnected regions of the parameter space. In this case, surrogate modelling and optimisation direction estimation are particularly challenging due to high uncertainty, thus making gradient-free optimisation much more preferable than gradient methods.

Common gradient-free optimisation methods, e.g., Bayesian optimisation, rely on uncertainty estimation over the parameter space to predict the potential minimum. We need to leverage the knowledge of the stochastic simulator response, hence an uncertainty estimator that would account for the complexity of both the response shape and the geometry of the objective function is required. We propose to minimise regret as in the lower confidence bound approach, but the variance of the predictive posterior is moved from $(\mathbb{R}^d, L_2)$ metric space to $(\mathcal{P}_2, \mathbb{W}_2)$ metric space. That is done by modelling the stochastic simulator response with a deep generative model (DGM) as the surrogate~\cite{calogan, Mustafa_Bard_Bhimji_Lukić_Al-Rfou_Kratochvil_2019}.

A well-trained DGM surrogate can properly replicate the simulator response. The use of DGMs as surrogates provides the opportunity to work in agnostic predictive posterior settings ~\cite{frazier2018tutorial}. In addition, a single run of a DGM surrogate is orders of magnitude faster than a simulator call, which is exactly the advantage of neural network surrogates over simulators.

We review related works in \hyperref[rel_work]{Section 2}. \hyperref[method]{Section 3} describes the proposed approach. In \hyperref[exp]{Section 4} we examine our approach on a set of experiments and compare our results with selected baselines. The results of the research are discussed and concluded in \hyperref[conclusion]{Section 5}. The appendices can be found in Ref.~\cite{ramazyan2024globaloptimisationblackboxfunctions}.

%%%%%%%%%%%%%%%%%%%%%%%%%%%%%%%%%%%%%%%%%%%%%%%%%%%%%%%%%%%%%%%%%%%%%%%%

\section{Related Work}
\label{rel_work}

Black-box optimisation encompasses numerous methods. Common methods include conjugate gradient \cite{conj_gradient}, Quasi-Newton methods, such as BFGS \cite{bfgs}, trust-region methods \cite{trust_region, trust_region_book}, and multi-armed bandit approaches \cite{bandits}. In particular, Ref. \cite{lgso, lgso10, lgso26, diffBBO} explore deep generative models for black-box optimisation. \cite{lgso10, lgso26} focus on the inference of posterior parameters using given observations, and \cite{lgso} uses direct gradient optimisation of the objective function using a local generative surrogate model. In addition, \cite{lgso20} and \cite{diffBBO} exploit the architecture of generative models used for optimisation - \cite{lgso20} optimises in a latent space that a DGM learns, and \cite{diffBBO} employ the invertibility of diffusion models to solve the inverse optimisation problem. Further, \cite{lgso18} aims to find optimal parameters of a DGM that approximate a fixed black box metric the best.

In practice, nondifferentiable objective functions are often needed. In this case Bayesian optimisation (BO)~\cite{lgso58, lgso55, bayes_opt}, genetic algorithms~\cite{lgso44, lgso5}, or numerical differentiation~\cite{lgso62} are preffered. BO is typically equipped with a Gaussian process (GP), which is defined by a mean function and a kernel function that describes the shape of the covariance. BO with a GP surrogate requires covariance matrix inversion with $O(n^3)$ cost in terms of the number of observations, which is challenging for a large number of responses and in higher dimensions. To make BO scalable \cite{lgso64, lgso16, lgso19, lgso66} consider a low-dimensional linear subspace and decompose it into subsets of dimensions, i.e., some structural assumptions are required that might not hold. In addition, GP requires a proper choice of kernel is crucial. BO may need the construction of new kernels \cite{lgso22}. \cite{duvenaud2014automatic} proposes a greedy approach for automatically building a kernel by combining basic kernels such as linear, exponential, periodic, etc., through kernel summation and multiplication.

BO is not bound to use GP. For example, deep Gaussian processes (DGP)~\cite{dgp, bo_dgp, bo_dgp_} attempt to resolve the issue of finding the best kernel for a GP. That is done by stacking GPs in a hierarchical structure as Perceptrons in a multilayer perceptron, but the number of variational parameters to be learnt by DGPs increases linearly with the number of data points, which is infeasible for stochastic black-box optimisation, and they have the same matrix inverting issue as GPs, which limits their scalability.

Other surrogate models that are reasonable to consider for BO are Bayesian neural networks (BNN) \cite{kononenko1989bayesian, bo_bnn} and ensemble models \cite{bo_de}. BNN directly quantify uncertainty in their predictions and generalise better than other neural networks. However, their convergence may be slower as each weight is taken from a distribution. This also means that BNN might require much more data than GPs to yield good predictions.

On the other hand, an intuitive approach for uncertainty quantification is using an ensemble surrogate model, e.g., an adversarial deep ensemble (DE) \cite{de, bo_de}. Single predictors of the ensemble are expected to agree on their predictions over observed regions of the feature space, i.e., where data are given and so the uncertainty is low and vice versa. The further these single predictors get from known regions of the feature space, the greater the discrepancy in their predictions.

We chose these surrogates as in this work, we focus on approaches that take advantage of the uncertainty incorporated in the predictive posteriors of the surrogate. GP, DGP, and BNN have a distribution as their outputs, and DE comprises one through its single predictors.

%%%%%%%%%%%%%%%%%%%%%%%%%%%%%%%%%%%%%%%%%%%%%%%%%%%%%%%%%%%%%%%%%%%%%%%%

\section{Method}
\label{method}

\subsection{Problem Statement}
\label{problem_statement}

In design optimisation, one aims to find a configuration of the system $\theta$ in the search space $\Theta \subseteq \mathbb{R}^m$, which minimises the expected cost $\mathbb{E}^{\mu} [f(\theta, x)]$ of the black box. $f$ is a real-valued function $f: \Theta \times \mathbb{R}^d \rightarrow \mathbb{R}$, where $x \sim \mu \in \mathcal{P}(\mathbb{R}^d)$, and $\mathcal{P}(\mathbb{R}^d)$ is the space of Borel probability measures in $\mathbb{R}^d$. This means that $f$ is pointwise observable, defined over a set of parameters $\Theta$. In this work, we take continuous $\Theta$. We assume that $f$ is bounded from below and Lipschitz continuous. The set $\Theta$ is convex, explicit, and non-relaxable, i.e., it does not rely on estimates and $f$ cannot be computed outside the feasible region. The assumption of finite first and second moments of $\mu$ is reasonable. It is denoted as $\mu \in \mathcal{P}_2$, where $\mathcal{P}_2$ is the set of probability measures with finite second moment.

As mentioned earlier, in black box optimisation simulated samples typically represent disjoint regions of the parameter space. This means that there is little or no information about most of the search space. In addition, the probability measure $\mu$ is unknown. Thus, following \cite{lanzetti2022firstorder}, we consider the problem of optimisation under uncertainty - a problem tackled via distributionally robust optimisation. The problem of uncertainty estimation amounts to evaluating the worst-case risk of the objective function over an ambiguity set $\mathcal{P} \subset \mathcal{P}_2(\mathbb{R}^d)$:

\begin{equation}
    \sup_{\mu \in \mathcal{P}} \mathbb{E}^{\mu} [f(\theta, x)].
\end{equation}

Generally, ambiguity sets are used to ensure statistical performance and computational feasibility \cite{ambiguity1}. In this work, we use the construction of considered ambiguity sets, Wasserstein balls to be precise, and the topology they induce to quantify uncertainty. The Wasserstein metric was studied for distributionally robust optimisation in Ref. \cite{ambiguity2, ambiguity3, wass_rob_opt}. %ambiguity4

Thus, we find the optimal configuration by solving the following optimisation problem:

\begin{equation} \label{obj}
    \inf_{\theta \in \Theta} \sup_{\mu \in \mathcal{P}} \mathbb{E}^{\mu} [f(\theta, x)].
\end{equation}

The optimal design is found by selecting a point of potential minimum among a set of explored candidates and calling the black box for that point. Consequently, filling the search space with ground truths and exploiting the obtained information increases the confidence of the predicted optimal configuration.

%%%%%%%%%%%%%%%%%%%%%%%%%%%%%%%%%%%%%%%%%%%%%%%%%%%%%%%%%%%%%%%%%%%%%%%%

\subsection{Deep Generative Models for Surrogate Modelling}
\label{dgms}

Having a generator model that would map a simple distribution, $\mathcal{Z}$, to a complex and unknown distribution, $p(\mu)$, is desired in many settings as it allows for the generation of samples from the intractable data space. For our purposes, conditional deep generative models (DGMs) are used to model the stochastic response of the simulator. This is particularly advantageous in exploring unknown regions of the parameter space. By parametrising the response space we exploit deep generative models in showing that they can be used for both approximating the objective function and modelling the simulator for the whole feature space and hence also computing the proposed uncertainty.

The goal of DGM is to obtain a generator $G: \mathbb{R}^s \rightarrow \mathbb{R}^d$ that maps a given distribution, e.g., a Gaussian, $\mathcal{Z}$ supported in $\mathbb{R}^s$ to a data distribution $p(\mu)$ supported in $\mathbb{R}^d$.

According to the problem statement, each sample is associated with a specific configuration, i.e., a vector of parameters $\theta$, so we consider conditional DGMs. Then the goal is to obtain $G: \mathbb{R}^s \times \Theta \rightarrow \mathbb{R}^d$ such that distributions $G(\mathcal{Z}; \theta)$ and $p(\mu; \theta)$ match for each $\theta \in \Theta$. Since $\mathcal{Z}$ and $p(\mu)$ are independent, we get $G(\mathcal{Z}; \theta) \sim p(\mu; \theta)$. For simplicity of notation we denote $G(\mathcal{Z}; \theta)$ as $G(\theta)$.

Thus, we explore the search space by training a deep generative surrogate $G$ on the set of ground truths $\mathcal{M}$. For each value $\theta \in \Theta$, we can generate the corresponding response of the simulator $\nu (\theta) = G(\theta)$, i.e., $\nu$ is the predictive distribution at $\theta$.

%%%%%%%%%%%%%%%%%%%%%%%%%%%%%%%%%%%%%%%%%%%%%%%%%%%%%%%%%%%%%%%%%%%%%%%%

\subsection{Wasserstein Uncertainty}
\label{wu}

In the usual Bayesian optimisation formulation, the epistemic uncertainty estimator is the variance of the predictive posterior distribution. Since the simulator is stochastic, we aim to have an uncertainty estimator that would account for the intricacies of both sample density and the geometry of the objective function. Hence, we propose the Wasserstein uncertainty - a Wasserstein distance-based uncertainty estimator that encapsulates the difference between known and predicted samples in both the distribution and the $L_2$ space.

\begin{definition}
    2-Wasserstein distance, denoted by $\mathbb{W}$, is the optimal transport cost between two probability distributions $\mu, \nu \in \mathcal{P}_2$ with $L_2^2$ cost function:

    \begin{equation}
        \mathbb{W}(\mu, \nu) = \left( \inf_{\pi \in \Pi(\mu, \nu)} \int |x-y|^2 d\pi(x, y) \right)^{\frac{1}{2}}.
    \end{equation}
    
    $\Pi(\mu, \nu)$ is the set of couplings of $\mu$ and $\nu$. The Wasserstein distance defines a symmetric metric on $\mathcal{P}_2$.
\end{definition}

Optimisation over an ambiguity set allows us to describe the regions of the parameter space with certain levels of uncertainty with respect to given ground truths. Ambiguity sets are usually defined to ensure statistical performance guarantees and computational tractability \cite{amb_sets1, amb_sets2}. A popular choice of ambiguity sets are Wasserstein balls \cite{wball1, wball2}.

\begin{definition}
    Wasserstein ball of radius $\varepsilon$ and centred at $\nu$, denoted by $\overline{B}_{\varepsilon}(\nu)$, is the closure with respect to the topology induced by the Wasserstein distance as follows:

    \begin{equation}
        \overline{B}_{\varepsilon}(\nu) := \{ \mu \in \mathcal{P}_2(\mathbb{R}^d): \mathbb{W}(\mu, \nu) \leq \varepsilon \}.
    \end{equation}
\end{definition}

This significantly eases the search for a potential optimum, as for a candidate $\nu$ its uncertainty amounts to:

\begin{equation} \label{eq:unc_1}
    F(\nu) = \sup_{\mu \in \overline{B}_{\varepsilon}(\nu)} \mathbb{E}^{\mu}[f].
\end{equation}

The uncertainty of $\nu$ is associated with the closest known response, i.e., for any prediction $\nu$ we consider $\Tilde{\mu} \in \mathcal{M}$ such that 
$\Tilde{\mu} = \arginf_{\mu \in \mathcal{M}} \mathbb{W}(\mu, \nu)$. From the uncertainty estimation perspective, we can be as certain about $\nu$ as about the closest to it $\Tilde{\mu} \in \mathcal{M}$. This also agrees with \cite{wass_rob_opt}, where a decision under uncertainty is modelled with a real-valued loss function, and the risk of the optimal decision is the least risky admissible loss.

Consider $\overline{B}_{\varepsilon}(\nu)$ a Wasserstein ball centred at $\nu$ with radius $\varepsilon = \mathbb{W}(\nu, \Tilde{\mu})$, i.e., such that $\Tilde{\mu}$ is located at the boundary of the ambiguity set. Considering any other radius of $\overline{B}_{\varepsilon}(\nu)$ would yield either overestimation or underestimation of the uncertainty of $\nu$.

In fact, for an approximation of $f$ of the form $\hat{f} = \langle w, x \rangle$, e.g., a neural approximator, the supremum is attained at the boundary of the Wasserstein ball.

\begin{lemma}
    (Existence of a minimum at the boundary, \cite{lanzetti2022firstorder}) There exists a worst-case probability measure $\mu^*$ attaining the supremum (\ref{eq:unc_1}) such that $\mathbb{W}(\nu, \mu^*) = \varepsilon$.
\end{lemma}

In general, finding $\mu^*$ is a task of its own, which would add computational costs to the optimisation problem. And $\nu$ would be equidistant from the closest known response, $\Tilde{\mu}$, and the worst-case risk of the objective function, $\mu^*$. Thus, for a predicted $\nu$, we quantify its uncertainty as:

\begin{equation} \label{eq:wu}
    \sigma_{\mathbb{W}} (\theta) = \inf_{\mu \in \mathcal{M}} \mathbb{W} (\mu, \nu).
\end{equation}

For $\sigma_{\mathbb{W}}$ to be a viable uncertainty estimator, it must be zero for known responses and vice versa. Lemma~\ref{lem:zero_w2} follows from the fact that $\mathbb{W}$ is a distance. The proof is provided in Appendix A ~\cite{ramazyan2024globaloptimisationblackboxfunctions}.

\begin{lemma}\label{lem:zero_w2}
    For a predicted random variable $\nu$ and a set of ground truths $\mathcal{M}$
    \begin{equation}
        \sigma_{\mathbb{W}}(\theta) = 0 \Leftrightarrow \exists \mu \in \mathcal{M}: \mu = \nu.
    \end{equation}
\end{lemma}

Calculating the true Wasserstein distance would add computational costs that would not marginally benefit the algorithm's performance. It is typically approximated with the corresponding $L_p$ norm. In all our experiments we consider only univariate real-valued random variables. According to Ref.~\cite{energy_dist}, $L_2$ and Energy distances are equivalent in this case. 

\begin{equation} \label{eq:energy}
\begin{split}
    D^2(\mu, \nu) = 2 \mathbb{E} \| X-Y\| - \mathbb{E}\|X-X'\| - \mathbb{E}\| Y-Y'\|,
\end{split}
\end{equation}

where $X, X' \sim F_{\mu}$, $Y, Y' \sim F_{\nu}$, and $F_{\mu}, F_{\nu}$ are the CDFs of $\mu$ and $\nu$ respectively. We estimate the Wasserstein uncertainty, Eq.~\ref{eq:wu}, for a predictive posterior $G(\theta)$ via Energy distance as follows:

\begin{equation}
    \hat{\sigma}_{\mathbb{W}}(\theta) = \min_{\mu \in \mathcal{M}} D(\mu, G(\theta)).
\end{equation}

% The 2-Wasserstein metic is studied for uncertainty estimation by \cite{wass_unc, w2_gp_unc}. \cite{w2_gp_unc} show the advantage of replacing naive averaging over GP predictions with the Wasserstein barycentre of said predictions. \cite{wass_unc} examines the quality of prediction of a stochastic response by a surrogate model that approximates the black-box function well enough. The problem has no solutions when posed in $L_p$ spaces, but in spaces induced with a Wasserstein metric we have upper and lower bounds on the predictions of the surrogate model.

%%%%%%%%%%%%%%%%%%%%%%%%%%%%%%%%%%%%%%%%%%%%%%%%%%%%%%%%%%%%%%%%%%%%%%%%

\subsection{Optimisation}
\label{opt}

Bayesian optimisation (BO) \cite{Mockus_1989} is perhaps the most common approach for global optimisation. It is based on acquisition functions. In BO the acquisition function is repeatedly applied until convergence.

The authors of the efficient global optimisation (EGO) algorithm~\cite{ego} proposed the expected improvement (EI) heuristic to fully exploit predictive uncertainty. The potential for optimisation is linked via a measure of improvement - a random variable defined for an input $\theta \in \Theta$ as

\begin{equation}
    I(\theta) = \max \{0, f_{min} - \nu(\theta) \},
\end{equation}

where $f_{min}$ is the best objective value obtained so far, and $\nu(\theta)$ is the predictive distribution of the fitted model at $\theta$. If $\nu(\theta)$ has a non-zero probability of taking any value on the real line, then $I(\theta)$ has a nonzero probability of being positive at any $\theta$.

To target potentials for large improvements more precisely, the EGO algorithm aims to maximise expected improvement $EI(\theta) = \mathbb{E}[I(\theta)]$. In predictive posterior agnostic settings EI is calculated using the MC approximation.

\begin{equation}
    EI(\theta) \approx \frac{1}{M} \sum_{j=1}^M \max \{ 0, f_{min} - x_j \},
\end{equation}

where $\{x_j\}_{j=1}^M, x_j \sim \nu(\theta)$. As $M \rightarrow \infty$ the approximation becomes exact. If $\nu(\theta)$ is Gaussian with mean $\mu(\theta)$ and variance $\sigma^2(\theta)$, e.g., as in Gaussian Processes, then EI accepts the following closed form.

\begin{equation} \label{eq:gaussian_ei}
% \begin{split}
    EI(\theta) = \sigma(\theta) \left[ z(\theta) \cdot \Phi\left( z(\theta) \right) + \phi\left( z(\theta) \right)\right],
% \end{split}
\end{equation}

where $z(\theta) = \frac{f_{min} - \mu(\theta)}{\sigma(\theta)}$ and $\Phi$ and $\phi$ are Gaussian CDF and PDF respectively.

Another common BO approach is lower confidence bound (LCB)~\cite{LAI19854}. Its acquisition function is regret defined as follows:

\begin{equation} \label{eq:regret_lcb}
    \mathcal{R}(\theta) = \mu(\theta) - \kappa \cdot \sigma(\theta).
\end{equation}

It is a linear combination of exploitation, $\mu(\theta)$, and exploration, $\sigma(\theta)$. The trade-off between the two is controlled via the hyperparameter $\kappa$. Smaller $\kappa$ yields more exploitation, and larger values of $\kappa$ yield more exploration of high-variance responses, where uncertainty is higher, i.e., where the black box is more unknown.

We omit any assumptions on the functional form of the posterior distribution and the computational and sampling challenges of Monte Carlo estimates. LCB can adjust exploration and exploitation, and EGO considers the predictive posterior distribution explicitly. We propose to combine both approaches and account for the stochasticity not explicitly in the objective as in EGO, but implicitly as in LCB. Thus, using the proposed in Eq. \ref{eq:wu} Wasserstein uncertainty, we suggest the following formulation of regret:

\begin{equation} \label{eq:wu_regret}
    \mathcal{R}_{\mathbb{W}} (\theta) = \mu(\theta) - \kappa \cdot \sigma_{\mathbb{W}}(\theta)
\end{equation}

This also agrees with the optimal solution of the supremum (\ref{eq:unc_1}) for a neural approximator (see Proposition 4.4 in Ref. \cite{lanzetti2022firstorder}).

Optimisation of acquisition functions such as EI can be done numerically. The proposed acquisition function, Eq.~\ref{eq:wu_regret}, is non-convex, similarly to LCB, and non-differentiable, which complicates its optimisation. We do it via direct search over a large set of candidate points. It simplifies the search for the next best set of parameters but limits the precision of the optimisation algorithm. For consistency of comparison, we use the direct search approach for all methods. This issue requires further studies.

%%%%%%%%%%%%%%%%%%%%%%%%%%%%%%%%%%%%%%%%%%%%%%%%%%%%%%%%%%%%%%%%%%%%%%%%

\subsection{Convergence}

% Convergence of gradient-free optimisation methods is not as straightforward as that of gradient approaches. The direction and the magnitude of a gradient-free optimisation step may change significantly from one iteration to another. This complicates the study of gradient-free optimisation convergence.

Under some theoretical assumptions, convergence rates of Bayesian optimisation can be guaranteed. The authors of \cite{ego_convergence} provide convergence rates for expected improvement algorithms. However, since in practice priors are estimated from data, standard estimators may not hold up to these convergence rates, and thus they need to be altered.

One may search for a more efficient method with improved average-case performance that still ensures reliable convergence. The challenge in developing such a method lies in balancing exploration and exploitation, as with LCB or WU-GO. Exploiting the data specifically in regions of the search space where the black-box function is known to be sub-optimal may help us find the optimum quickly. However, without exploring all regions of the search space, we might never find the optimal design or be confident in the optimality of the proposed solution \cite{convergence}. This is further discussed in the ablation study reported in Appendix C ~\cite{ramazyan2024globaloptimisationblackboxfunctions}.

In general, to converge to the exact solution, as with any other gradient-free optimisation approach, it may take an infinite number of simulator calls. One might deduce a worst-case scenario, but it may require an unfeasibly large number of simulator calls. Since such scenarios are unlikely in practice and the number of simulations is limited, it makes sense to relax the guarantees to address practical constraints while still feasible solutions.

% \begin{theorem}\label{thm} For a parameter space $\Theta \subset \mathbb{R}^d$ given by a hypercube of diameter $\ell$ that contains the optimum of an $L$-Lipschitz stochastic black-box function $f$, the number of iterations required to convergence to an $\varepsilon$-solution is:

% \begin{equation}
%      n \leq 2 ^ {d \cdot log_{\frac{1}{\varepsilon}} \left( \ell \right)},
% \end{equation}

% and in general the diameter $\ell$ can be replaced with $L \cdot \mathbb{W}_1(\mu_0, \mu_1)$, where $\mathbb{W}_1$ is the 1-Wasserstein metric, and $\mu_0$ and $\mu_1$ lie on the ends of $\Theta$'s diameter.

% \end{theorem}

% % In general, to converge to the exact solution, as with any other gradient-free optimisation approach, it may take an infinite amount of simulator calls. 

% We deduce a worst-case scenario for WU-GO in Theorem~\ref{thm}. It is unlikely to happen in practice and it may suggest an unfeasibly large number of simulator calls. Since the number of simulators is typically limited, it makes sense to relax the guarantees to address practical constraints while still providing feasible solutions.

%%%%%%%%%%%%%%%%%%%%%%%%%%%%%%%%%%%%%%%%%%%%%%%%%%%%%%%%%%%%%%%%%%%%%%%%

\subsection{Algorithm}

\begin{algorithm}[ht] 
   \caption{Wasserstein Uncertainty Global Optimisation (WU-GO)}\label{alg:wu_alg}
\begin{algorithmic}
   \STATE {\bfseries Input:} Ground truths $\mathcal{M}$, generator $G$, candidates $\Tilde{\Theta}$, parameter $\kappa$
   \STATE {\bfseries Output:} Optimal configuration $\hat{\theta}^*$
   \\\hrulefill
   \WHILE{stopping criteria are not met}
   \STATE Fit $G$ on $\mathcal{M}$
   % \STATE Generate $\Phi = \{\phi_{\theta}^{(j)} \sim G(\theta); j \in [n]; \forall \theta \in \Tilde{\Theta}\}$
   \STATE Approximate $f$: $\hat{f}(\theta) = \mathbb{E}[G(\theta)] \approx \frac{1}{n} \sum_{j=1}^n x_j, x_j \sim G(\theta)$
   \STATE Estimate $\sigma_{\mathbb{W}}$: $\hat{\sigma}_{\mathbb{W}}(\theta) = \min_{\mu \in \mathcal{M}} D(\mu, G(\theta))$
   \STATE Predict $\hat{\theta} = \argmin_{\theta \in \Tilde{\Theta}} \{ \hat{f}(\theta) - \kappa \cdot \hat{\sigma}_{\mathbb{W}}(\theta) \} \}$
   \STATE Call simulator for $\hat{\theta}: \hat{\mu}$
   \STATE $\mathcal{M} = \mathcal{M} \cup \{ \hat{\mu} \}$
   \ENDWHILE
\end{algorithmic}
\end{algorithm}

The proposed algorithm is summarised in Algorithm \ref{alg:wu_alg}. Our algorithm requires a generative surrogate model and chooses the next point minimising $\mathcal{R}_{\mathbb{W}}$ from Eq. \ref{eq:wu_regret}.

As stated in Algorithm \ref{alg:wu_alg}, the black-box function is evaluated over a set of candidate values represented by a grid $\Tilde{\Theta} \subset \Theta$ representing the search space. As discussed in Section 3.4, This is a numerical limitation, as the Wasserstein uncertainty is not differentiable. On the other hand, this allows us to easily consider configurations and subspaces of interest, and define stopping criteria, i.e., an $\varepsilon$-solution criterion.

To select the value of $\kappa$, we conduct an ablation study. The value $\kappa = 2$ is the best among the considered ones (see Appendix C ~\cite{ramazyan2024globaloptimisationblackboxfunctions} for more details).

%%%%%%%%%%%%%%%%%%%%%%%%%%%%%%%%%%%%%%%%%%%%%%%%%%%%%%%%%%%%%%%%%%%%%%%%

\subsection{Surrogate Model and Baselines} \label{sec:models}

WU-GO is not bound to any specific generative surrogate model. In this work, we take a conditional WGAN-GP \cite{gulrajani2017improved}. Using GANs is advantageous for faster sampling, i.e., faster exploration of the search space. Gradient penalty provides a better objective function approximation, and the Wasserstein uncertainty could be directly estimated from the surrogate model, which would be highly efficient, especially for multidimensional outputs. However, \cite{exact_wass} and \cite{wgans_fail} argue that for the true Wasserstein distance, one should alter the training of the WGAN.

As baselines, we take EGO and LCB with Gaussian posterior. As surrogates for these algorithms, we take Gaussian process regression, Deep Gaussian Processes, Bayesian neural networks, and adversarial deep ensembles. An effort was made to make all models as similar as possible to achieve more consistent results. Appendix B ~\cite{ramazyan2024globaloptimisationblackboxfunctions} contains model parameters and experiment settings for reproducibility.

%%%%%%%%%%%%%%%%%%%%%%%%%%%%%%%%%%%%%%%%%%%%%%%%%%%%%%%%%%%%%%%%%%%%%%%%

\section{Experiments}
\label{exp}

We evaluate WU-GO performance using eight experiments. A real working example of a high energy physics detector and seven experiments based on commonly used optimisation objective functions. In each synthetic test, we change the settings of the experiment to examine the models in various scenarios. A short description is provided below and for a full description see Appendix D ~\cite{ramazyan2024globaloptimisationblackboxfunctions}.

\begin{table}[h]
\caption{Experiments.} \label{table:exp}
\vskip 0.15in
\begin{center}
\begin{small}
\begin{sc}
\begin{tabular}{lccc}
\toprule
Black Box Function   & $N$   & $n$    & $\Theta$        \\%& $|\Tilde{\Theta}|$\\
\midrule
Three Hump Camel     & $4$   & $10^2$ & $[-5, 5]^2$     \\%& $101^2$ \\
Ackley               & $4$   & $10^2$ & $[-5, 5]^2$     \\%& $101^2$ \\
Lévi                 & $4$   & $10^2$ & $[-4, 6]^2$     \\%& $101^2$ \\
Himmelblau           & $4$   & $10^1$ & $[-5, 5]^2$     \\%& $101^2$ \\
Rosenbrock           & $121$ & $10^2$ & $[-2, 2]^8$     \\%& $101^2$ \\
Rosenbrock           & $25$  & $10^2$ & $[-2, 2]^{20}$  \\%& $101^2$ \\
Styblinki-Tang       & $25$  & $10^2$ & $[-5, 5]^{20}$  \\%& $101^2$ \\
\bottomrule
\end{tabular}
\end{sc}
\end{small}
\end{center}
\vskip -0.1in
\end{table}

\begin{figure*}[ht]
\centering
\includegraphics[width=\linewidth]{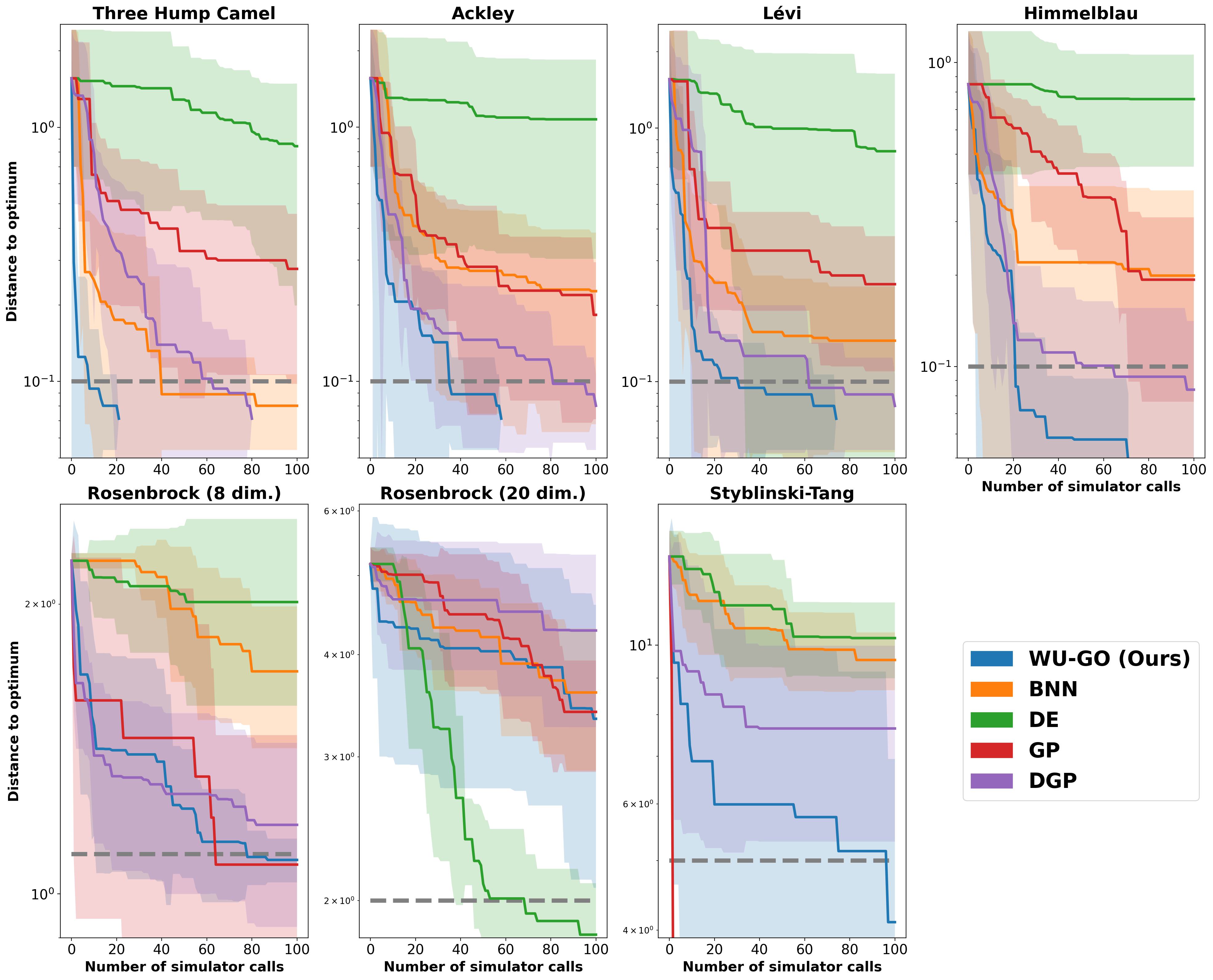}
\caption{Experiment results - comparison of EGO and WU-GO. See Section 4.1 for more details.}
\label{fig:res_plots}
\end{figure*}

Similarly to experiments of ~\cite{audet2023sequential}, all instances are stochastic. Each experiment is defined by the black-box response shape. Namely, $\mathcal{N}(f(\theta), 10^{-2})$, where $f(\theta)$ is the black-box function. The only exception is the Lévi experiment, where the variance response is a bimodal Gaussian changing the shape with the function.

Table~\ref{table:exp} provides the parameters for each experiment. For each test function, we conduct $10$ runs for $N$ different starting configurations chosen via Latin hypercube sampling. Each response is represented by a sample of size $n$. Optimisation is performed in the search space $\Theta$ among a set of candidate points $\Tilde{\Theta}$ consisting of $101^2$ points. %The dimensionality of synthetic experiments ranges from $2$ to $20$, and the physics experiment search space is 22-dimensional, thus covering a range of 

% For each test function, we conduct 10 runs for different starting configurations $\hat{\Theta} \subset \Theta, |\hat{\Theta}| = N$ - the initial positioning of $\mu \in \mathcal{M}$ is different at each run. For two-dimensional parameter spaces, i.e., the Three Hump Camel, Ackley, Lévi, and Himmelblau experiments, we select $N=4$ points and $N=121$ points for the 8-dimensional Rosenbrock test using orthogonal Latin hypercube sampling. For 20-dimensional search spaces, we start with $N=25$ points that we choose with Latin hypercube sampling. The sample size of each response $\mu$ is $n=10^2$. Except the Himmelblau experiment, for which we take a smaller sample size to test in the scenario when the black box response is underrepresented.

% As in tests we use functions with known minima, using Proposition 3 we choose an epsilon as the stopping criteria for the tests. The epsilon is chosen as the distance to one or several best possible minima. In addition to that, as most test functions are polynomials we apply log transformation to the black box function to ease surrogate model training.

%%%%%%%%%%%%%%%%%%%%%%%%%%%%%%%%%%%%%%%%%%%%%%%%%%%%%%%%%%%%%%%%%%%%%%%%

\subsection{Results}
\label{results}

We evaluate all models in terms of distance to the optimum and speed of convergence. As each iteration of an optimisation algorithm would require a call of the simulator, we measure the speed of convergence as the number of simulator calls. In Fig.~\ref{fig:res_plots} we report the mean and the standard deviation of the distance to minimum w.r.t. the number of simulator calls.

Table~\ref{table:prob} and Table~\ref{table:dist} provide insights on results from Fig.~\ref{fig:res_plots} in a cross-sectional manner. Table~\ref{table:prob} showcases the probability of a model to yield an $\varepsilon$-solution within the first $100$ iterations. That is done by considering each model result as a Bernoulli random variable with the probability of success being the proportion of runs when the model met the stopping criteria. Table~\ref{table:dist} is a numerical representation of Fig.~\ref{fig:res_plots} - it shows the distance to optima after $50$ simulator calls. Top-3 results for each experiment are highlighted in green, yellow, and red respectively.

\begin{table*}[t]
\caption{Probability to yield an $\varepsilon$-solution within the first 100 iterations.} \label{table:prob}
\vskip 0.15in
\begin{center}
\begin{small}
\begin{sc}
\begin{tabular}{lccccc}
\toprule
Black Box Function   & WU-GO                                 & BNN                                    & DE                                      & GP                 & DGP                \\ \midrule
Three Hump Camel     & \cellcolor{green!25}$1.0 \pm 0.000$  & \cellcolor{yellow!25}$0.9 \pm 0.095$ & $0.1 \pm 0.095$    & \cellcolor{red!25}$0.2 \pm 0.126$ & \cellcolor{green!25}$1.0 \pm 0.000$ \\
Ackley               & \cellcolor{green!25}$1.0 \pm 0.000$  & $0.2 \pm 0.126$                      & $0.1 \pm 0.095$     & \cellcolor{red!25}$0.4 \pm 0.155$ & \cellcolor{yellow!25}$0.9 \pm 0.095$ \\
Lévi                 & \cellcolor{green!25}$1.0 \pm 0.000$  & \cellcolor{red!25}$0.5 \pm 0.158$    & $0.0 \pm 0.000$    & $0.2 \pm 0.126$ & \cellcolor{yellow!25}$0.9 \pm 0.095$ \\
Himmelblau           & \cellcolor{green!25}$1.0 \pm 0.000$  & \cellcolor{red!25}$0.4 \pm 0.155$    & $0.0 \pm 0.000$   & $0.2 \pm 0.126$ & \cellcolor{yellow!25}$0.7 \pm 0.145$  \\
Rosenbrock (8 dim.)  & \cellcolor{yellow!25}$0.6 \pm 0.155$ & $0.0 \pm 0.000$                      & $0.1 \pm 0.095$    & \cellcolor{green!25}$0.9 \pm 0.095$ & \cellcolor{red!25}$0.5 \pm 0.158$ \\
Rosenbrock (20 dim.) & \cellcolor{yellow!25}$0.3 \pm 0.145$ & $0.0 \pm 0.000$                    & \cellcolor{green!25}$0.8 \pm 0.126$ & $0.0 \pm 0.000$              & $0.0 \pm 0.000$              \\
Styblinki-Tang       & \cellcolor{yellow!25}$0.8 \pm 0.126$ & $0.0 \pm 0.000$                  & $0.0 \pm 0.000$                                   & \cellcolor{green!25}$1.00 \pm 0.000$  & \cellcolor{red!25}$0.3 \pm 0.145$   \\
\bottomrule
\end{tabular}
\end{sc}
\end{small}
\end{center}
\vskip -0.1in
\end{table*}

\begin{table*}[t]
\caption{Distance to optimum after 50 simulator calls.} \label{table:dist}
\vskip 0.15in
\begin{center}
\begin{small}
\begin{sc}
\begin{tabular}{lccccc}
\toprule
Black Box Function   & WU-GO                                 & BNN                                    & DE                                      & GP                 & DGP                \\ \midrule
Three Hump Camel     & \cellcolor{green!25}$0.071 \pm 0.000$  & \cellcolor{yellow!25}$0.089 \pm 0.035$  & $1.286 \pm 0.796$    & $0.326 \pm 0.240$ & \cellcolor{red!25}$0.131 \pm 0.151$ \\
Ackley               & \cellcolor{green!25}$0.089 \pm 0.035$ & \cellcolor{red!25}$0.272 \pm 0.198$   & $1.109 \pm 0.779$    & $0.283 \pm 0.209$ & \cellcolor{yellow!25}$0.146 \pm 0.091$ \\
Lévi                 & \cellcolor{green!25}$0.089 \pm 0.035$  & \cellcolor{red!25}$0.157 \pm 0.105$  & $0.996 \pm 0.978$   & $0.329 \pm 0.138$ & \cellcolor{yellow!25}$0.126 \pm 0.074$ \\
Himmelblau           & \cellcolor{green!25}$0.058 \pm 0.034$ & \cellcolor{red!25}$0.220 \pm 0.172$   & $0.760 \pm 0.307$   & $0.397 \pm 0.180$ & \cellcolor{yellow!25}$0.103 \pm 0.063$  \\
Rosenbrock (8 dim.)  & \cellcolor{green!25}$1.226 \pm 0.188$ & \cellcolor{red!25}$1.977 \pm 0.317$ & $2.048 \pm 0.363$    & $1.452 \pm 0.624$ & \cellcolor{yellow!25}$1.270 \pm 0.270$ \\
Rosenbrock (20 dim.) & \cellcolor{yellow!25}$3.934 \pm 1.460$ & \cellcolor{red!25}$4.205 \pm 0.623$  & \cellcolor{green!25}$2.210 \pm 0.398$   & $4.483 \pm 0.827$ & \cellcolor{yellow!25}$4.667 \pm 0.837$ \\
Styblinki-Tang       & \cellcolor{yellow!25}$5.993 \pm 4.452$  & $10.490 \pm 1.479$  & $11.233 \pm 1.924$   & \cellcolor{green!25}$2.067 \pm 0.028$  & \cellcolor{red!25}$7.651 \pm 2.334$   \\
\bottomrule
\end{tabular}
\end{sc}
\end{small}
\end{center}
\vskip -0.1in
\end{table*}

In the main body of this paper we compare WU-GO with EGO with baseline surrogates, as described in Section~\ref{sec:models}. The discrepancy between WU-GO and EGO results is less evident and should be discussed. The baselines with the LCB approach performed significantly worse. We place LCB results in Appendix E ~\cite{ramazyan2024globaloptimisationblackboxfunctions}.

A faster decrease of mean values in Fig.~\ref{fig:res_plots} implies faster convergence of the algorithm. Smaller values of standard deviation mean higher confidence in predictions, i.e., consistency of the mean value. A combination of the two implies the efficiency of the model - not only does the model converge but it is also confident in its predictions.

In experiments with two-dimensional search spaces, WU-GO attains minima more efficiently than all selected baselines. That is not the case for, e.g., DE in the Lévi experiment or BNN in the Himmelblau experiment, in which standard deviation spreads over a large portion of the presented domain. Among the baselines, the performance of DGP is closest to that of WU-GO.
%in the sense that the mean results converge and the standard deviation decreases as the model converges.

The Himmelblau test shows that despite responses being under-represented - only 10 observations were simulated for each sample - WU-GO manages to outperform all baselines. It is likely that it would not be able to reproduce the original densities precisely. Still, the black-box function approximation is good enough for the optimisation task.

This is all evidence that WU-GO carries out optimisation efficiently of complex functions, e.g., Ackley and Lévi, as well as it does on simpler tests such as the Three Hump Camel. This signals that WU-GO should be more robust in general than EGO. It is worth noting that the results across all four two-dimensional experiments are similar. This shows the consistency of all models.

As the dimensionality of the search space increases, convergence is expected to decrease. We see this effect with WU-GO. However, the only time DE converges and shows a stable performance is in the 20-dimensional Rosenbrock experiment. GP behaves similarly. Although GP converges in the 8-dimensional Rosenbrock and the 20-dimensional Styblicky-Tang experiment, they fail with the 20-dimensional Rosenbrock. EGO may marginally outperform WU-GO, but the performance varies significantly depending on the surrogate model. This should make one question the stability and the consistency of EGO.

Although WU-GO's performance worsens, it converges in over half of the runs and is the only other model to show such results except for the previously mentioned DE and GP on specific tests. WU-GO's performance can be improved by tuning $\kappa$. It does require additional computations, but the resulting performance may be worth it as in the two-dimensional experiments (see ablation study in Appendix C ~\cite{ramazyan2024globaloptimisationblackboxfunctions} for more details).

\subsection{Physics experiment}

The muon shield is an essential element of the Search for Hidden Particles (SHiP) experiment, deflecting the abundant muon flux generated within the target away from the detector, otherwise a substantial background for particle searches. Muon deflection out of the spectrometer's acceptance by a magnetic field is straightforward; however, the challenge lies in the wide distribution of muons in phase space. To address this, SHiP uses both magnetic and passive shielding, as detailed in Ref.~\cite{Akmete_2017}, to safeguard the emulsion target from muons.

To refine the muon shielding and optimize its configuration, a GEANT4 \cite{geant} simulation was developed to track muons' trajectories through the magnets. Each magnet's specifications encompass seven parameters: length, widths at both ends, heights, and air-gap widths between the field and return field. Overall we consider 22 parameters describing the architecture and the geometry of the detector. Thus we aim to improve the muon shield cost, $\mathbb{E}[f]$, over the space of experiment responses, $\mu \in \mathcal{P}_2$, by altering the configuration of the muon shield, $\theta \in \Theta$.

\begin{figure}[h]
\centering
\includegraphics[width=\linewidth]{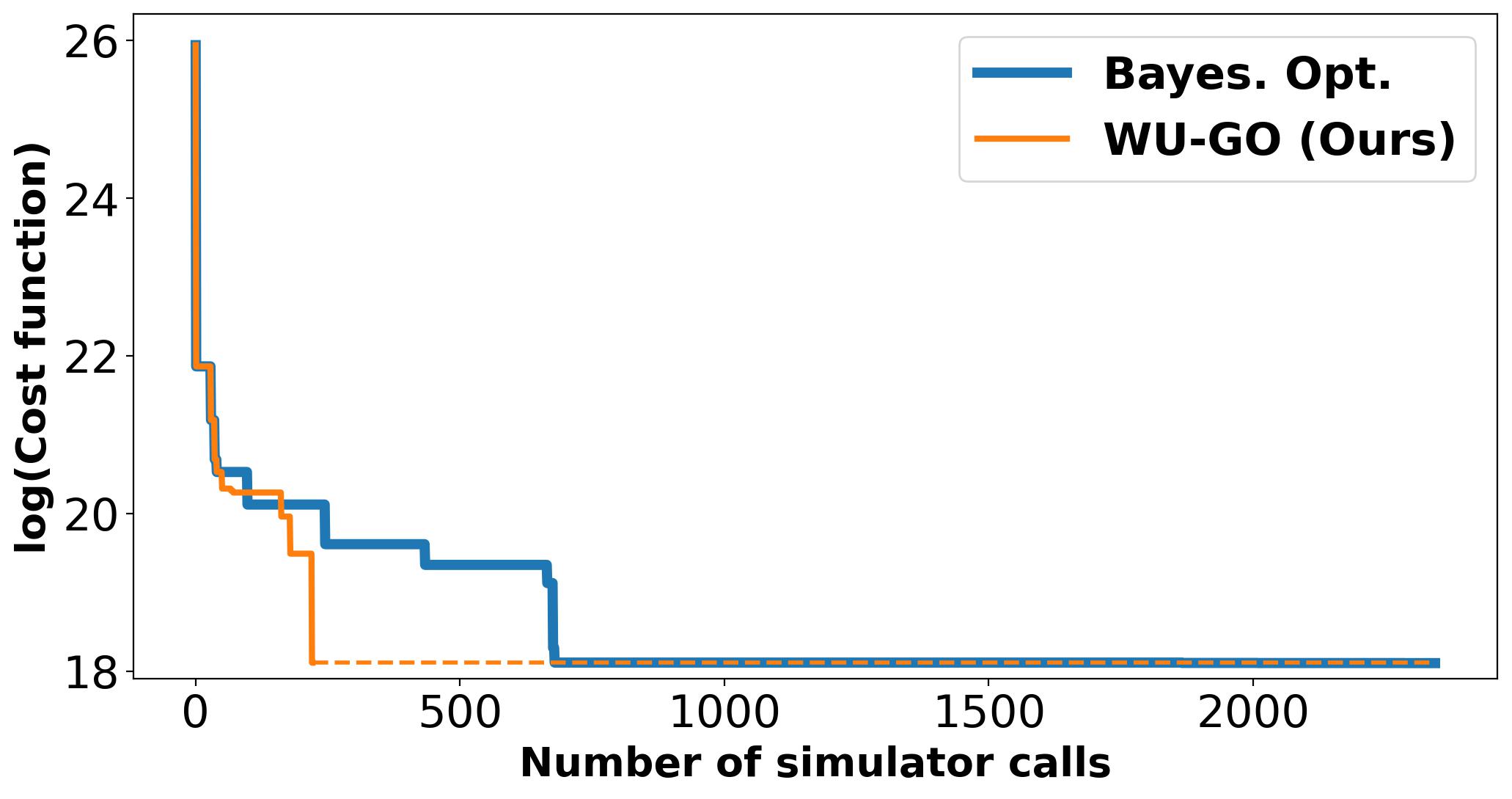}
\caption{SHiP Muon Shield optimisation results.}
\label{fig:ship_res}
\end{figure}

\vspace{0.5cm}

The objective function of the muon shield optimisation is a combination of shielding efficiency and a mass cost function. We compare the result of WU-GO with a previously obtained result by Bayesian optimisation with a Gaussian process surrogate. The design suggested via Bayesian optimisation was approved for the construction of the experiment.

Within the first $300$ iterations, WU-GO suggested a configuration of the SHiP experiment with almost identical performance as that suggested by Bayesian optimisation. In contrast with Bayesian optimisation, WU-GO also suggested some lightweight and compact designs. Although the shielding efficiency is suboptimal, the smaller dimensions of the muon shield may allow us to combine the muon shield with the rest of the experiment in a more efficient manner, i.e., minimise the background noise. This will be the subject of a future work of ours.

The starting configuration is described in Ref.~\cite{Akmete_2017} and all above-mentioned muon shield configurations can be found in Appendix F ~\cite{ramazyan2024globaloptimisationblackboxfunctions}.

% For instance, a general pipeline of engineering optimisation consists of several steps~\cite{pipeline}. Each configuration under consideration must be analysed using a sufficiently large simulated dataset and a well-tuned parameter reconstruction procedure. Possible parameters range from the size of the setup to the inner composition of the material, i.e., a non-differentiable dimension of the parameter space. The choice of configuration is carried out using a surrogate model of the optimised system based on the reconstruction of the simulated sample.

It is worth noting that the Bayesian optimisation stops after $2500$ iterations because it runs out of memory. WU-GO is constant in memory. Hence we are not as limited in the number of optimisation loop iterations and we may gather as much statistics as we would like. In fields such as high energy physics, this feature is significant.

Bayesian optimisation with a GP surrogate model may be the most common approach to a design optimisation problem. With a GP surrogate, the following two issues may also arise.

The first challenge occurs when ground-truth observations cover the search space with large gaps in between. As previously mentioned, this is a general occurrence in black-box optimisation. Especially, when the search space is of higher dimensionality, e.g., the 22-dimensional search space of SHiP experiment parameters. In this case, GP tends to underestimate the uncertainty within the gaps of the search space. 

The second issue arises when the variance of ground-truth observations depends on the derivative of the black-box function. Then the GP variance estimates tend to the variance of all ground truth samples, i.e., some average value. Thus the variance is generally mispredicted.

Both are illustrated on toy examples in Appendix G ~\cite{ramazyan2024globaloptimisationblackboxfunctions}. The GP is fitted well in the sense that the black-box function approximation is close. However, the aforementioned issues of mispredicted variance may yield mispredictions of both EGO and LCB optimisation algorithms.

%%%%%%%%%%%%%%%%%%%%%%%%%%%%%%%%%%%%%%%%%%%%%%%%%%%%%%%%%%%%%%%%%%%%%%%%

% \section{Broader Impact}
% \label{discussion}

% %%%%%%%%%%%%%%%%%%%%%%%%%%%%%%%%%%%%%%%%%%%%%%%%%%%%%%%%%%%%%%%%%%%%%%%%

\section{Conclusion}
\label{conclusion}

We propose a novel approach for gradient-free optimisation of stochastic non-differentiable simulators. The algorithm is based on the concept of Wasserstein balls as ambiguity sets for uncertainty quantification. It uses a deep generative surrogate model, which allows us to model and optimise entities from simple Gaussian random variables to more complicated cases. We perform experiments on a real high energy physics detector simulator and a series of toy problems covering a wide range of possible real experiment settings. Our method, WU-GO, is compared with efficient global optimisation and lower confidence bound with Bayesian neural network, adversarial deep ensemble, Gaussian process, and deep Gaussian process as surrogate models. WU-GO attains minima more efficiently in comparison with all selected baselines, with DGP results being the closest - typically worse by a margin of one standard deviation. Experiments also suggest that our approach is robust to the shape of both the response and the objective function. WU-GO is easily extendible to higher dimensions of the parameters space but might require tuning of the exploration v. exploitation hyperparameter $\kappa$. Our method may not just mitigate issues of commonly used approaches but resolve them - a deep generative surrogate model can handle high-dimensional inputs, yield highly accurate reconstruction for various required configurations, and generate large samples with low computing time. WU-GO may be a universal approach for optimisation problems, especially for non-differentiable objectives and when reconstruction plays a key role in the optimisation pipeline.

\section*{Acknowledgements}

The research leading to these results has received funding from the Basic Research Programme at the National Research University Higher School of Economics. This research was supported in part through computational resources of HPC facilities at HSE University.

%%%%%%%%%%%%%%%%%%%%%%%%%%%%%%%%%%%%%%%%%%%%%%%%%%%%%%%%%%%%%%%%%%%%%%%%

%%% Use this command to include your bibliography file.

\bibliography{mybibfile}

\newpage

%%%%%%%%%%%%%%%%%%%%%%%%%%%%%%%%%%%%%%%%%%%%%%%%%%%%%%%%%%%%%%%%%%%%%%

\appendix
\onecolumn

\section*{Appendix A. Proofs} \label{proofs}

\textbf{Lemma 2.} For a generated random variable $\nu$ and a set of ground truths $\mathcal{M}$

\begin{equation}
    \sigma_{\mathbb{W}}(\theta) = 0 \Leftrightarrow \exists \mu \in \mathcal{M}: \mu = \nu.
\end{equation}

\begin{proof}
    Let's start by proving the equivalence from left to right.
    \begin{equation}
        \sigma_{\mathbb{W}}(\theta) = 0 \Leftrightarrow \inf_{\mu \in \mathcal{M}} \mathbb{W}(\mu, \nu) = 0 \Rightarrow \exists \mu \in \mathcal{M}: \mathbb{W}(\mu, \nu) = 0.
    \end{equation}
    As $\mathbb{W}_2$ is a distance, from the statement above we have that $\phi = \psi$.
    
    Now, let's prove the equivalence from right to left.
    
    If $\mu = \nu$, then $\mathbb{W}(\mu, \nu) = 0$. As $\mathbb{W}$ is non-negative, $\inf_{\mu \in \mathcal{M}} \mathbb{W}(\mu, \nu) = 0 \Leftrightarrow \sigma_{\mathbb{W}}(\theta) = 0$. Hence $\exists \mu \in \mathcal{M}$ such that $\sigma_{\mathbb{W}}(\theta) = 0$. \newline
\end{proof}

\newpage

\section*{Appendix B. Models and Reproducibility} \label{appendix_models}

\begin{itemize}
    \item \textbf{WU-GO}:
        \begin{itemize}
            \item GAN with both Generator and Discriminator architecture: Linear - Tanh - Linear;
            \begin{itemize}
                \item latent dimension: $10$;
                \item hidden size: $64$;
                \item gradient penalty: $\lambda_{GP}= 1$;
                \item number of discriminator iterations per generator iteration: $5$;
                \item Adam optimiser, learning rate: $10^{-3}$;
                \item number of epochs: $100$.
            \end{itemize}
            \item Scheduler starts at the second iteration of the optimisation algorithm:
            \begin{itemize}
                \item scheduler: StepLR;
                \item starting learning rate: $10^{-1}$;
                \item multiplicative factor of learning rate decay: $\gamma = 10^{-1}$;
                \item period of learning rate decay: $30$ epochs;
            \end{itemize}
            \item Python library of implementation: \href{https://pytorch.org}{PyTorch}.
    \end{itemize}
    \item \textbf{BNN}:
        \begin{itemize}
            \item architecture: Bayes Linear - Tanh - Bayes Linear;
            \item mean of prior normal distribution of Bayes Linear: $\mu = 0$;
            \item sigma of prior normal distribution of Bayes Linear: $\sigma = 10^{-1}$;
            \item hidden size: $64$;
            \item KL weight: $10^{-2}$;
            \item number of predictions to average over: $10$;
            \item Python library of implementation: \href{https://bayesian-neural-network-pytorch.readthedocs.io/en/latest/}{Torch BNN}.
        \end{itemize}
    \item \textbf{DE}:
        \begin{itemize}
            \item Single regressor:
            \begin{itemize}
                \item architecture: Linear - Tanh - Linear;
                \item hidden size: $16$;
            \end{itemize}
            \item number of single regressors: $10$;
            \item Adam optimiser;
            \item scheduler: StepLR;
            \item starting learning rate: $10^{-1}$;
            \item multiplicative factor of learning rate decay: $\gamma = 10^{-1}$;
            \item period of learning rate decay: $6$ epochs;
            \item Python library of implementation: \href{https://ensemble-pytorch.readthedocs.io}{Ensemble PyTorch}.
        \end{itemize}
    \item \textbf{GP}:
        \begin{itemize}
            \item RBF kernel;
            \item Python library of implementation: \href{https://gpy.readthedocs.io}{GPy}.
        \end{itemize}
    \item \textbf{DGP}:
        \begin{itemize}
            \item architecture: RBF kernel - RBF kernel;
            \item hidden size: $10$;
            \item Python library of implementation: \href{https://github.com/SheffieldML/PyDeepGP}{PyDeepGP}.
        \end{itemize}
\end{itemize}

As mentioned in Section 3.7, we try to keep all models as similar to each other as possible as long as it does not hurt model performance.

%%%%%%%%%%%%%%%%%%%%%%%%%%%%%%%%%%%%%%%%%%%%%%%%%%%%%%%%%%%%%%%%%%%%%%%%%%%%%%%%%%%%%%%%%%
%%%%%%%%%%%%%%%%%%%%%%%%%%%%%%%%%%%%%%%%%%%%%%%%%%%%%%%%%%%%%%%%%%%%%%%%%%%%%%%%%%%%%%%%%%
%%%%%%%%%%%%%%%%%%%%%%%%%%%%%%%%%%%%%%%%%%%%%%%%%%%%%%%%%%%%%%%%%%%%%%%%%%%%%%%%%%%%%%%%%%

\newpage

\section*{Appendix C. Ablation study} \label{ablation}

\begin{figure}[!ht]
\centering
\includegraphics[width=\linewidth]{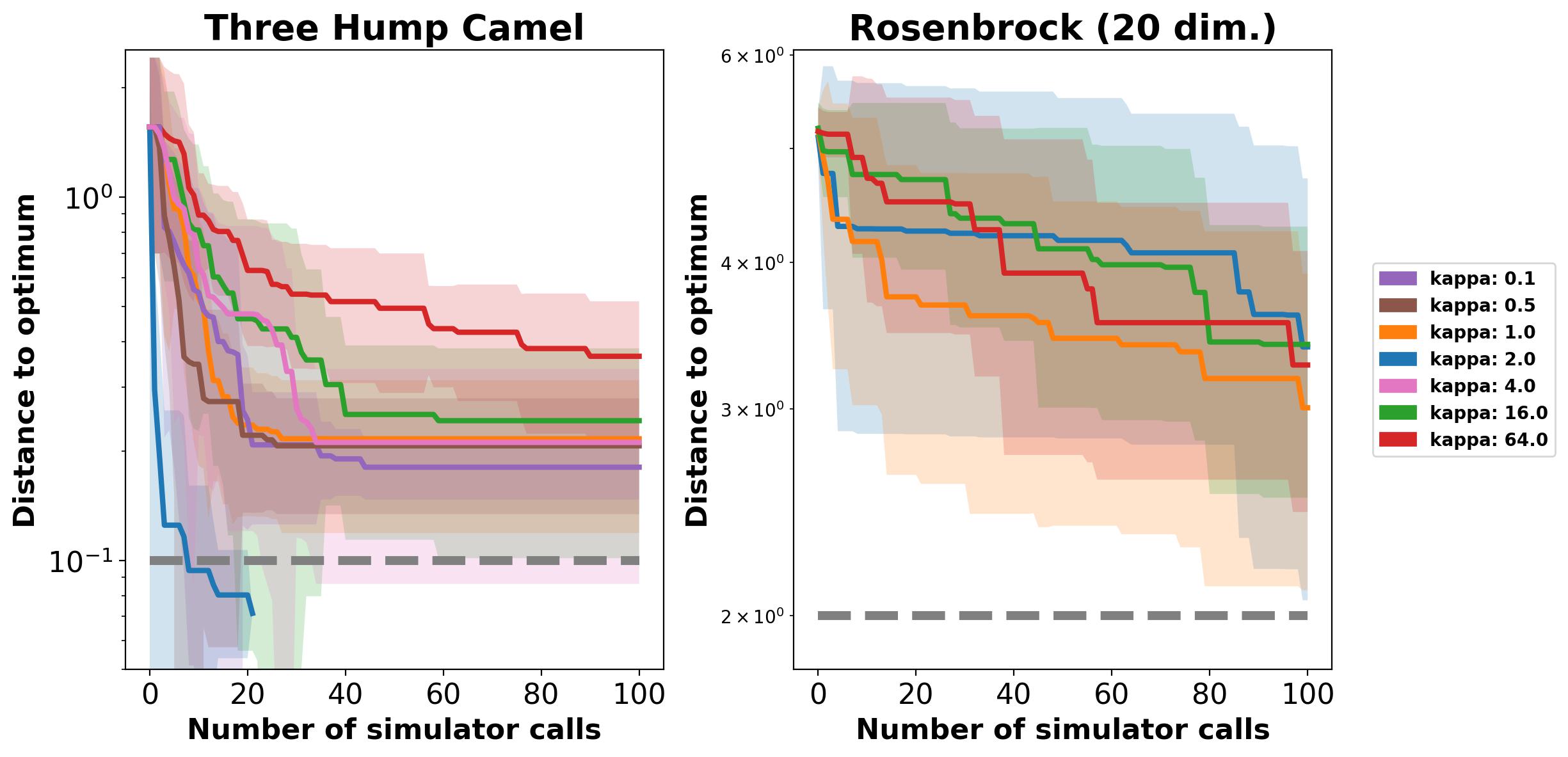}
\caption{Ablation study for optimal values of $\kappa$.}
\label{fig:ablation}
\end{figure}

\vspace{0.5cm}

In contrast with the EGO algorithm, WU-GO, similarly to the LCB approach, has a hyperparameter controlling the exploration-exploitation tradeoff of the algorithm - $\kappa$. The hyperparameter is similar to the learning rate in a gradient-based optimisation. Convergence of WU-GO highly depends on $\kappa$. The optimality of a specific $\kappa$ is contingent on the number of starting points and the dimensionality of the search space $\Theta$.

We study the performance of WU-GO for different values of $\kappa$ - we run the Three Hump Camel (2-dimensional search space) and the Rosenbrock (20-dimensional search space) experiments for a range of values of $\kappa$. Fig ~\ref{fig:ablation} shows the results of the ablation study.

In the Three Hump Camel experiment we see that $\kappa = 2.0$ is significantly better than any other tested value. As WU-GO with other $\kappa$'s does not even converge. Thus for all the above-described experiments, we had fixed the values of $\kappa = 2.0$.

One would assume that with an increase in the search space dimensionality, one ought to consider larger values of $\kappa$. However, the ablation study on the 20-dimensional Rosenbrock shows the opposite. With the smaller $\kappa = 1.0$, WU-GO converges better than with the default $\kappa = 2.0$. On the other hand, larger values, e.g., $\kappa = 16.0$ and $\kappa=64.0$, also improve the previously obtained result starting from the fortieth iteration of the algorithm.

As Fig.~\ref{fig:res_plots} and Fig.~\ref{fig:ablation} suggest, randomly selecting the $\kappa$ coefficient may result in suboptimal predictions at best. Ideally, it should be tuned for each configuration of an experiment. This does come with the cost of additional tests to run. But, as described in results obtained from numerical experiments, it may be worth it as WU-GO outperformed EGO in most experiments. \newline

The optimality of the $\kappa$ hyperparameter depends not only on the number of points initially given to the model and the dimensionality of the search space but also on their relative position. This question should and will be investigated in our future works.

%%%%%%%%%%%%%%%%%%%%%%%%%%%%%%%%%%%%%%%%%%%%%%%%%%%%%%%%%%%%%%%%%%%%%%%%%%%%%%%%%%%%%%%%%%
%%%%%%%%%%%%%%%%%%%%%%%%%%%%%%%%%%%%%%%%%%%%%%%%%%%%%%%%%%%%%%%%%%%%%%%%%%%%%%%%%%%%%%%%%%
%%%%%%%%%%%%%%%%%%%%%%%%%%%%%%%%%%%%%%%%%%%%%%%%%%%%%%%%%%%%%%%%%%%%%%%%%%%%%%%%%%%%%%%%%%

\newpage

\section*{Appendix D. Experiments full description} \label{lcb}

\textbf{Three Hump Camel} is a base experiment with a polynomial black-box function and the following response:

\begin{equation}
    \mu \sim \mathcal{N}\left( \mu; f(\theta), 10^{-2} \right) \text{ s.t. } f(\theta) = 2\theta_1^2 - 1.05\theta_1^4 + \frac{1}{6}\theta_1^6 + \theta_1\theta_2 + \theta_2^2
\end{equation}
\begin{center}
    in the search space $\Theta = [-5, 5]^2$ with global optimum $\theta^* = (0, 0), f(\theta^*) = 0.$
\end{center}

\textbf{Ackley} addresses the case of a complex black-box function with multiple local optima:
\begin{center}
    $\mu \sim \mathcal{N}\left( \mu; f(\theta), 10^{-2} \right)$ s.t.
\end{center}
\begin{equation}
    f(\theta) = -20 \exp \left[ -\frac{1}{5} \sqrt{\frac{1}{2}(\theta_1^2 + \theta_2^2)} \right] - \exp \left[ \frac{1}{2}\left( \cos(2 \pi \theta_1) + \cos(2 \pi \theta_2) \right) \right] + e + 20
\end{equation}
\begin{center}
    in the search space $\Theta = [-5, 5]^2$ with global optimum $\theta^* = (0, 0), f(\theta^*) = 0.$
\end{center}

\textbf{Lévi} addresses the case of response changing shape correspondingly to the black box function:
\begin{center}
    $\mu \sim \mathcal{N}\left( \mu; f(\theta), \sigma_{\beta}^2 \right)$ s.t. 
\end{center}
\begin{center}
    $\sigma_{\beta}^2 = \frac{1}{100} (\beta [4 - 3 \sin^2(3\pi\theta_2)] + (1 - \beta)[0.1 + 3 \sin^2(3\pi\theta_2)] ), \beta \sim Bern(n, 0.5)$ and
\end{center}
\begin{equation}
    f(\theta) = \sin^2(3\pi\theta_1) + (\theta_1 - 1)^2 (1 + \sin^2(3\pi\theta_2)) + (\theta_2 - 1)^2 (1 + \sin^2(2\pi\theta_2))
\end{equation}
\begin{center}
    in the search space $\Theta = [-4, 6]^2$ with global optimum $\theta^* = (1, 1), f(\theta^*) = 0.$
\end{center}

\textbf{Himmelblau} considers an underrepresented response, i.e., the sample size is decreased to $n=10$:

\begin{equation}
    \mu \sim \mathcal{N}\left( \mu; f(\theta), 10^{-2} \right) \text{ s.t. } f(\theta) = (\theta_1^2 + \theta_2 - 11)^2 + (\theta_1 + \theta_2^2 - 7)^2
\end{equation}
\begin{center}
    in the search space $\Theta = [-5, 5]^2$ with global optima $\theta^* = \begin{cases} (-3.0, -2.0) \\ (-2.8005118, 3.131312) \\ (-3.779310, -3.283186) \\ (3.584428, -1.848126) \end{cases}, f(\theta^*) = 0.$
\end{center}

\textbf{Rosenbrock} is a higher dimensional experiment - tested for dimensions $m=8$ and $m=20$ and significantly denser starting configurations for the 8-dimensional test:

\begin{equation}
    \mu \sim \mathcal{N}\left( \mu; f(\theta), 10^{-2} \right) \text{ s.t. } f(\theta) = \sum_{i=1}^{q-1} \left[ 100(\theta_{i+1} - \theta_i^2) + (1 - \theta_i)^2 \right]
\end{equation}
\begin{center}
    in the search space $\Theta = [-2, 2]^{q}$ with global optimum $\theta^* = \mathbf{1}_{q}, f(\theta^*) = 0.$
\end{center}

\textbf{Styblinski-Tang} is a higher-dimensional experiment $(m=20)$ with a polynomial black-box function:

\begin{equation}
    \mu \sim \mathcal{N}\left( \mu; f(\theta), 10^{-2} \right) \text{ s.t. } f(\theta) = \frac{1}{2} \sum_{i=1}^{q} \theta_i^4 - 16 \theta_i^2 + 5 \theta_i
\end{equation}

\begin{center}
    in the search space $\Theta = [-5, 5]^{q},$
\end{center}
\begin{center}
    with global optimum $\theta^* = (\! \underbrace{-2.903534, \cdots -2.903534\,}_\text{$q$ times}), -39.16617 \cdot q < f(\theta^*) < -39.16616 \cdot q.$
\end{center}

%%%%%%%%%%%%%%%%%%%%%%%%%%%%%%%%%%%%%%%%%%%%%%%%%%%%%%%%%%%%%%%%%%%%%%%%%%%%%%%%%%%%%%%%%%
%%%%%%%%%%%%%%%%%%%%%%%%%%%%%%%%%%%%%%%%%%%%%%%%%%%%%%%%%%%%%%%%%%%%%%%%%%%%%%%%%%%%%%%%%%
%%%%%%%%%%%%%%%%%%%%%%%%%%%%%%%%%%%%%%%%%%%%%%%%%%%%%%%%%%%%%%%%%%%%%%%%%%%%%%%%%%%%%%%%%%

\newpage

\section*{Appendix E. Lower Confidence Bound Results} \label{lcb}

\begin{figure*}[ht]\label{fig:plots}
\centering
\includegraphics[width=0.75\linewidth]{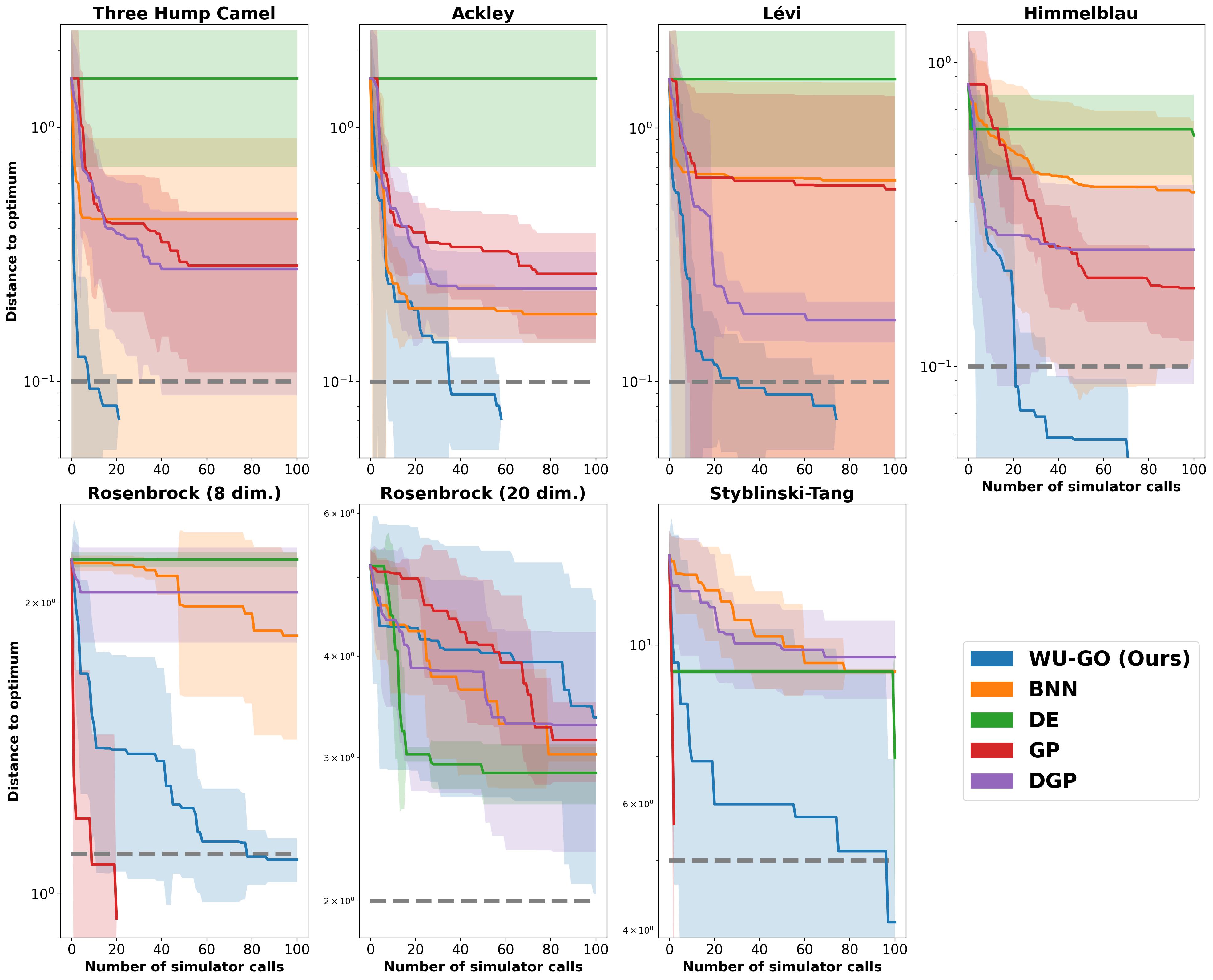}
\caption{Experiment results - comparison of LCB and WU-GO.}
\end{figure*}

As mentioned in Section 4.1 lower confidence bound (LCB) baselines yield results that may look similar to EGO ones. However, the performance of LCB baselines is significantly worse. Results presented in this appendix are reported exactly as in Section 4.1 only with baseline models using the LCB algorithm for optimisation. Both Fig.~\ref{fig:plots} and Table~\ref{table:dist_lcb} agree with the aforementioned that efficient global optimisation and LCB results look similar. However, Table~\ref{table:prob_lcb} reveals that except for Bayesian neural networks and deep ensembles in some runs and Gaussian processes converging in all runs in the 8-dimensional Rosenbrock and the Styblinki-Tang experiments, all baseline surrogate models with LCB do not converge.

\begin{table*}[ht]
\caption{Probability to yield an $\varepsilon$-solution within the first 100 iterations.} \label{table:prob_lcb}
\vskip 0.15in
\begin{center}
\begin{small}
\begin{sc}
\begin{tabular}{lccccc}
\toprule
Black Box Function   & WU-GO                                 & BNN                                    & DE                                      & GP                 & DGP                \\ \midrule
Three Hump Camel     & \cellcolor{green!25}$1.0 \pm 0.000$  & $0.0 \pm 0.000$ & $0.0 \pm 0.000$    & $0.0 \pm 0.000$ & $0.0 \pm 0.000$ \\
Ackley               & \cellcolor{green!25}$1.0 \pm 0.000$  & $0.0 \pm 0.000$ & $0.0 \pm 0.000$    & $0.0 \pm 0.000$ & $0.0 \pm 0.000$ \\
Lévi                 & \cellcolor{green!25}$1.0 \pm 0.000$  & $0.0 \pm 0.000$ & $0.0 \pm 0.000$    & $0.0 \pm 0.000$ & $0.0 \pm 0.000$ \\
Himmelblau           & \cellcolor{green!25}$1.0 \pm 0.000$  & $0.0 \pm 0.000$ & $0.0 \pm 0.000$   & $0.0 \pm 0.000$ & $0.0 \pm 0.000$  \\
Rosenbrock (8 dim.)  & \cellcolor{yellow!25}$0.6 \pm 0.155$ & \cellcolor{red!25}$0.1 \pm 0.095$ & $0.0 \pm 0.000$    & \cellcolor{green!25}$1.0 \pm 0.000$ & $0.0 \pm 0.000$ \\
Rosenbrock (20 dim.) & \cellcolor{green!25}$0.3 \pm 0.145$ & $0.0 \pm 0.000$ & $0.0 \pm 0.000$    & $0.0 \pm 0.000$  & $0.0 \pm 0.000$              \\
Styblinki-Tang       & \cellcolor{yellow!25}$0.8 \pm 0.126$ & $0.0 \pm 0.000$ & \cellcolor{red!25}$0.2 \pm 0.126$   & \cellcolor{green!25}$1.00 \pm 0.000$  & $0.0 \pm 0.000$   \\
\bottomrule
\end{tabular}
\end{sc}
\end{small}
\end{center}
\vskip -0.1in
\end{table*}

\begin{table*}[ht]
\caption{Distance to optimum after 50 simulator calls.} \label{table:dist_lcb}
\vskip 0.15in
\begin{center}
\begin{small}
\begin{sc}
\begin{tabular}{lccccc}
\toprule
Black Box Function   & WU-GO                                 & BNN                                    & DE                                      & GP                 & DGP                \\ \midrule
Three Hump Camel     & \cellcolor{green!25}$0.071 \pm 0.000$  & $0.436 \pm 0.473$  & $1.560 \pm 0.860$    & \cellcolor{red!25}$0.295 \pm 0.173$ & \cellcolor{yellow!25}$0.277 \pm 0.189$ \\
Ackley               & \cellcolor{green!25}$0.089 \pm 0.035$ & \cellcolor{yellow!25}$0.194 \pm 0.047$   & $1.560 \pm 0.860$    & $0.339 \pm 0.129$ & \cellcolor{red!25}$0.232 \pm 0.091$ \\
Lévi                 & \cellcolor{green!25}$0.089 \pm 0.035$  & $0.636 \pm 0.884$  & $1.560 \pm 0.860$   & \cellcolor{red!25}$0.619 \pm 0.744$ & \cellcolor{yellow!25}$0.185 \pm 0.040$ \\
Himmelblau           & \cellcolor{green!25}$0.058 \pm 0.034$ & $0.397 \pm 0.319$   & $0.604 \pm 0.178$   & \cellcolor{yellow!25}$0.214 \pm 0.063$ & \cellcolor{red!25}$0.242 \pm 0.154$  \\
Rosenbrock (8 dim.)  & \cellcolor{yellow!25}$1.226 \pm 0.188$ & \cellcolor{red!25}$1.991 \pm 0.386$ & $2.218 \pm 0.041$    & \cellcolor{green!25}$0.943 \pm 0.000$ & $2.051 \pm 0.232$ \\
Rosenbrock (20 dim.) & $3.934 \pm 1.460$ & \cellcolor{yellow!25}$3.640 \pm 0.780$  & \cellcolor{green!25}$2.945 \pm 0.200$   & $4.133 \pm 0.839$ & \cellcolor{red!25}$3.829 \pm 1.127$ \\
Styblinki-Tang       & \cellcolor{yellow!25}$5.993 \pm 4.452$  & $10.291 \pm 1.607$  & \cellcolor{red!25}$9.190 \pm 0.085$   & \cellcolor{green!25}$5.628 \pm 3.562$  & $10.007 \pm 1.394$   \\
\bottomrule
\end{tabular}
\end{sc}
\end{small}
\end{center}
\vskip -0.1in
\end{table*}

\newpage

\section*{Appendix F. SHiP muon shield configurations} \label{ship_configs}

\begin{figure}[!ht]
\centering
\includegraphics[width=\linewidth]{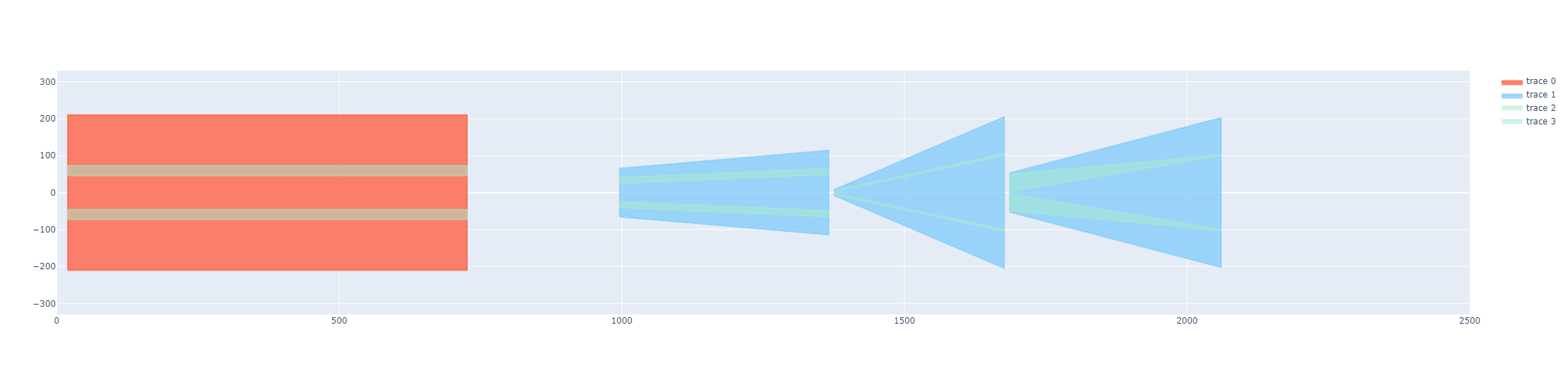}
\caption{Starting SHiP muon shield configuration.}
\label{fig:ship_default}
\end{figure}

\begin{figure}[!ht]
\centering
\includegraphics[width=\linewidth]{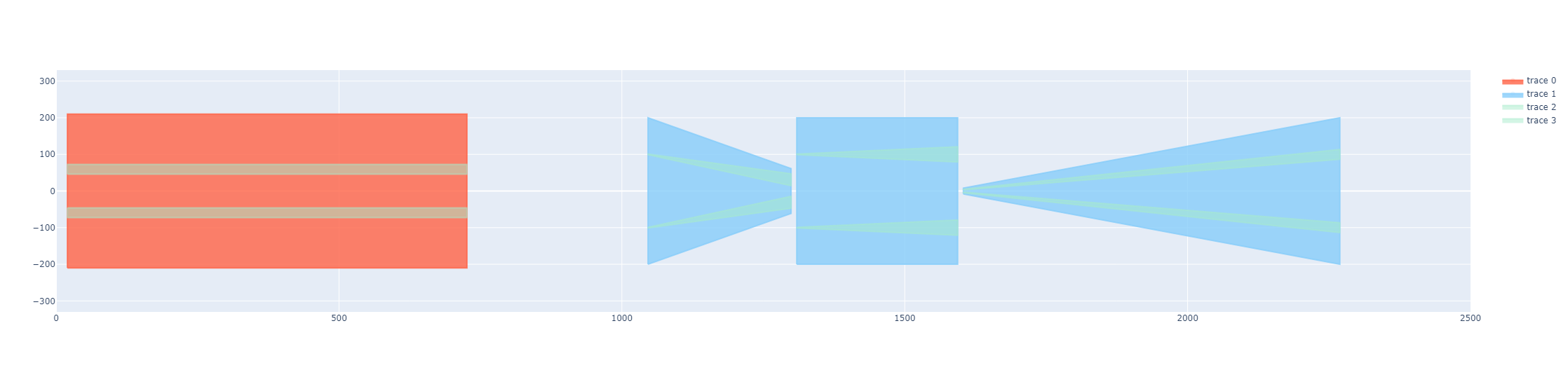}
\caption{SHiP muon shield configuration suggested by Bayesian optimisation.}
\label{fig:ship_bo}
\end{figure}

\begin{figure}[!ht]
\centering
\includegraphics[width=\linewidth]{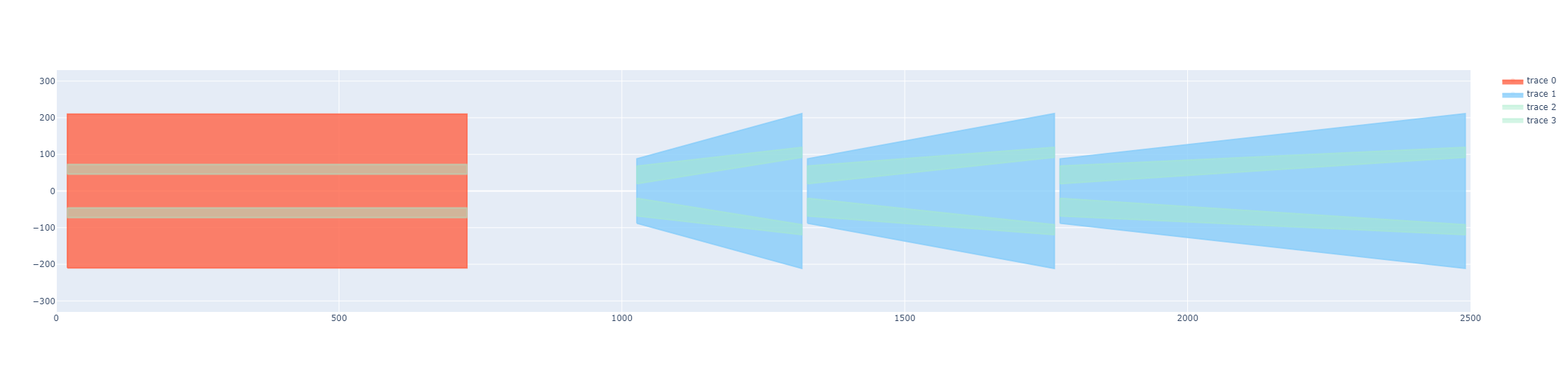}
\caption{SHiP muon shield configuration suggested by WU-GO.}
\label{fig:ship_wugo}
\end{figure}

\begin{figure}[!ht]
\centering
\includegraphics[width=\linewidth]{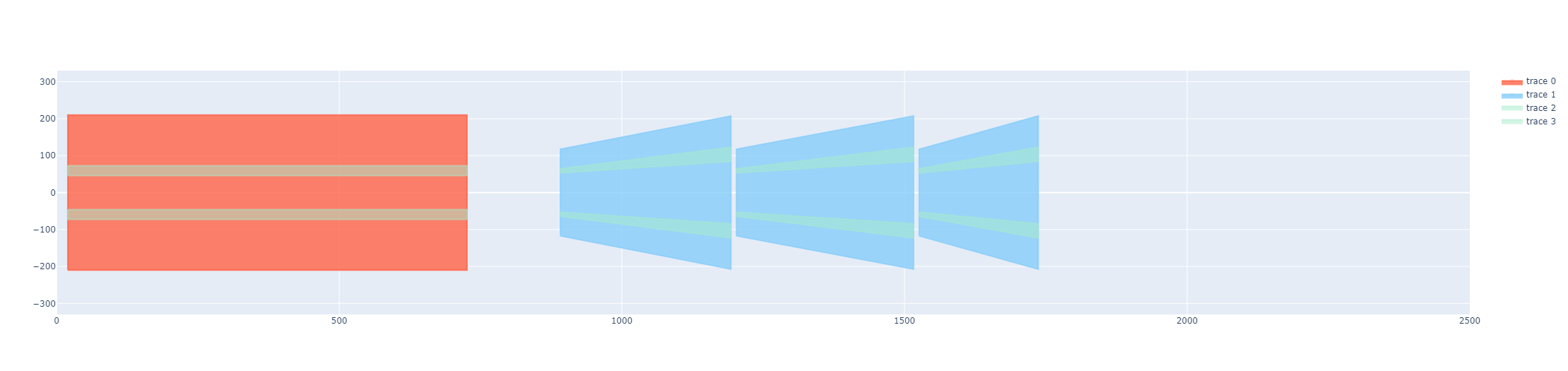}
\caption{Lightweight and compact SHiP muon shield configuration suggested by WU-GO.}
\label{fig:ship_wugo_small}
\end{figure}

\newpage

\section*{Appendix G. Gaussian process surrogate variance misprediction cases} \label{ship_configs}

\begin{figure}[!ht]
\centering
\includegraphics[width=\linewidth]{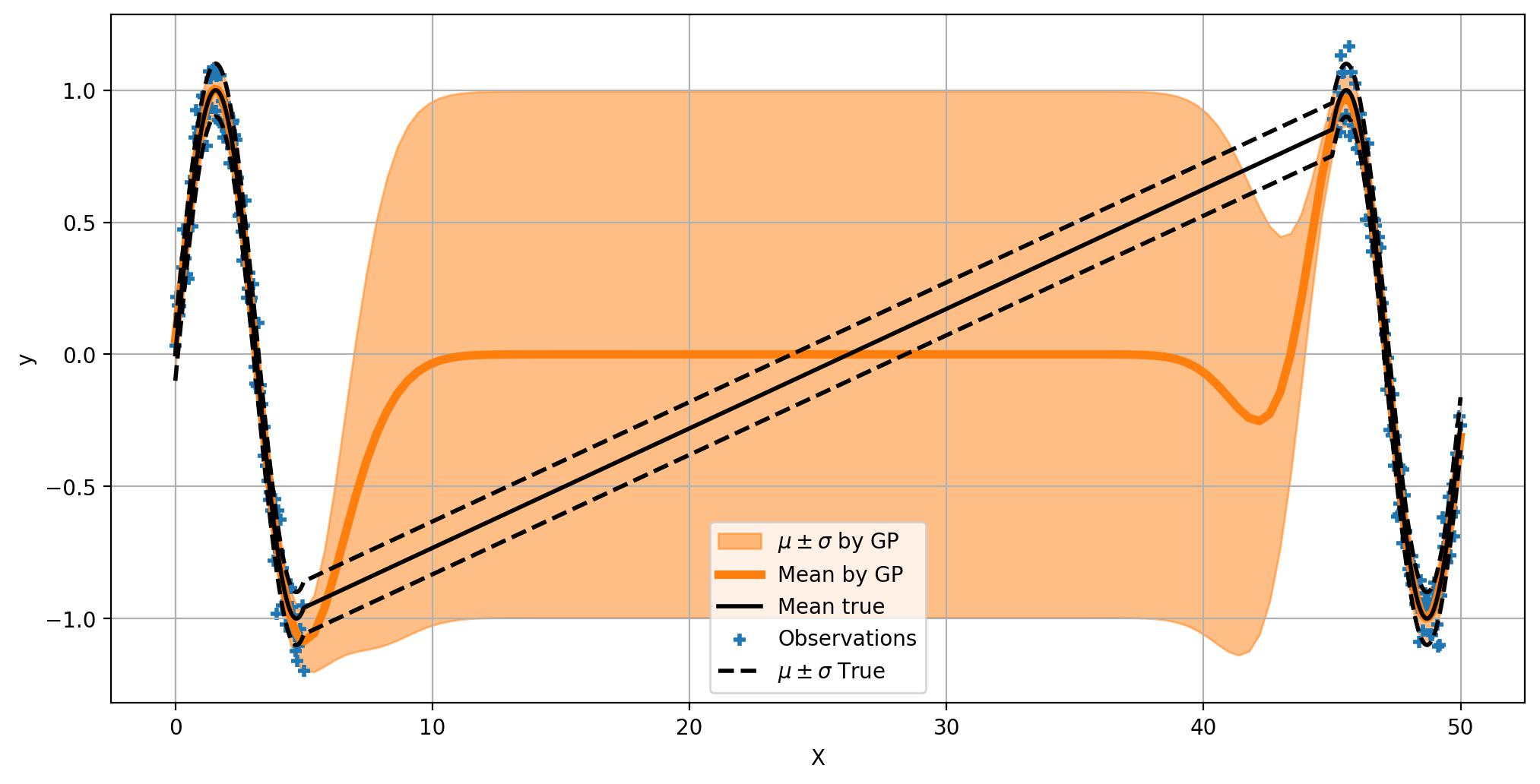}
\includegraphics[width=\linewidth]{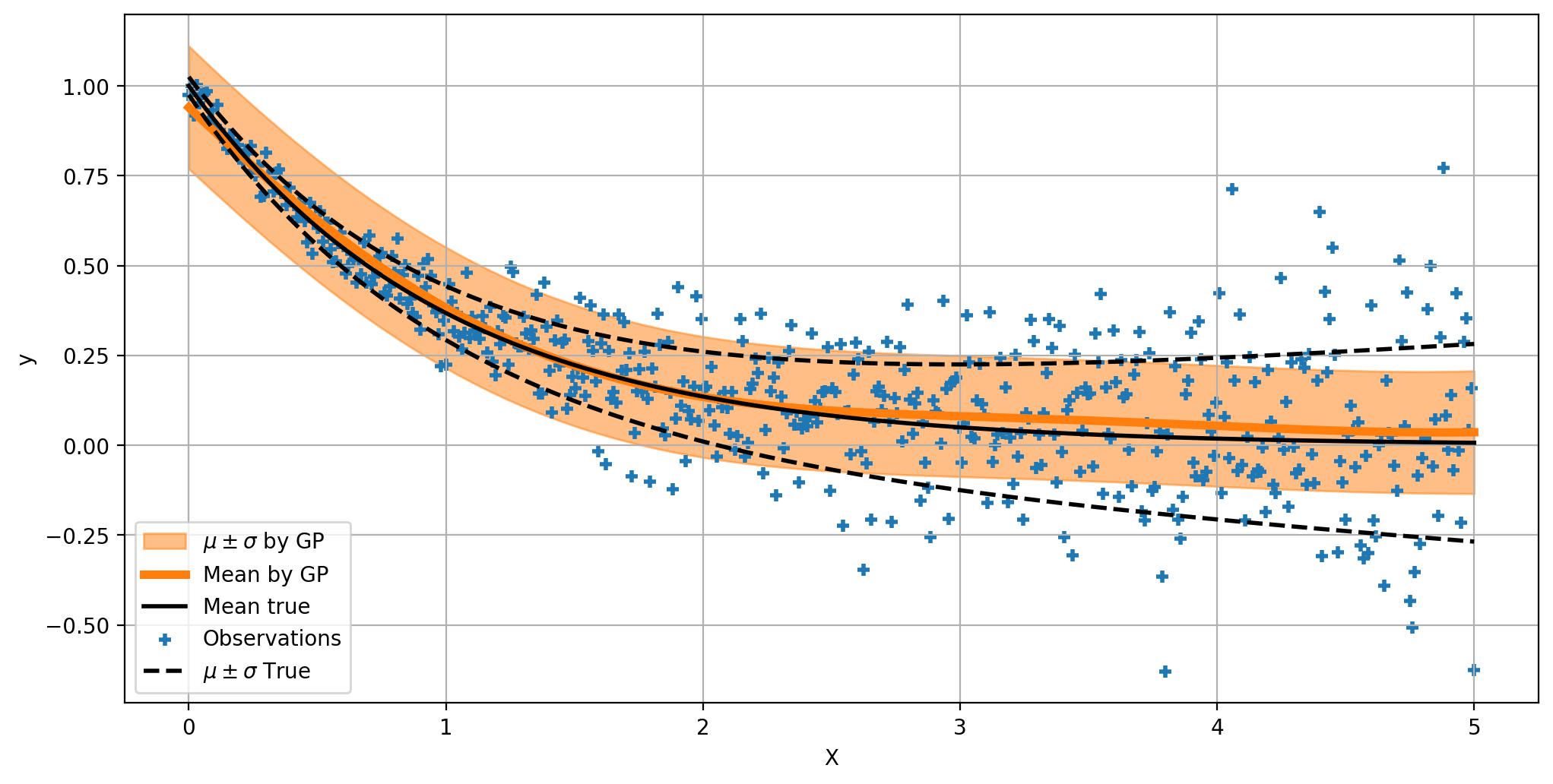}
\caption{Examples of GP surrogates mispredicting variance.}
\label{fig:gp}
\end{figure}

\end{document}